\documentclass[letterpaper, 10 pt, conference]{ieeeconf}
\IEEEoverridecommandlockouts 
\usepackage{cite}
\usepackage{amsmath,amssymb,amsfonts}
\usepackage{algorithmic}
\usepackage[algo2e]{algorithm2e} 
\usepackage{graphicx}
\usepackage{textcomp}
\let\labelindent\relax
\usepackage[inline]{enumitem}
\usepackage{xcolor}
\usepackage{caption}
\usepackage{graphicx}
\usepackage{algorithm}  
\usepackage{subfigure}
\usepackage[T1]{fontenc}
\usepackage{aecompl}

\usepackage{diagbox,slashbox}
\usepackage{hyperref}
\usepackage{verbatim}
\usepackage{dsfont}
\def\BibTeX{{\rm B\kern-.05em{\sc i\kern-.025em b}\kern-.08em
    T\kern-.1667em\lower.7ex\hbox{E}\kern-.125emX}}

\newenvironment{myitemize}{\begin{list}{$\bullet$}
{\setlength{\topsep}{1mm}
\setlength{\itemsep}{0.25mm}
\setlength{\parsep}{0.25mm}
\setlength{\itemindent}{0mm}
\setlength{\partopsep}{0mm}
\setlength{\labelwidth}{15mm}
\setlength{\leftmargin}{4mm}}}{\end{list}}

\begin{document}


\title{\bf Learning Representation for Anomaly Detection of Vehicle Trajectories\\
}


\author{Ruochen Jiao$^{1}$, Juyang Bai$^{1}$, Xiangguo Liu$^{1}$, Takami Sato$^{2}$, Xiaowei Yuan$^{1}$, Qi Alfred Chen$^{2}$, Qi Zhu$^{1}$%
\thanks{$^{1}$ Ruochen Jiao, Juyang Bai, Xiangguo Liu, Xiaowei Yuan, and Qi Zhu are with the Department of Electrical and Computer Engineering, Northwestern University, IL, USA.}%
\thanks{$^{2}$ Takami  Sato and Qi Alfred Chen are with the Department of Computer Science, University of California, Irvine, CA, USA.}%
}



\maketitle

\begin{abstract}
Predicting the future trajectories of surrounding vehicles based on their history trajectories is a critical task in autonomous driving. However, when small crafted perturbations are introduced to those history trajectories, the resulting anomalous (or adversarial) trajectories can significantly mislead the future trajectory prediction module of the ego vehicle, which may result in unsafe planning and even fatal accidents. Therefore, it is of great importance to detect such anomalous trajectories of the surrounding vehicles for system safety,  but few works have addressed this issue. 
In this work, we propose two novel methods for learning effective and efficient representations for online anomaly detection of vehicle trajectories.  Different from general time-series anomaly detection, anomalous vehicle trajectory detection deals with much richer contexts on the road and fewer observable patterns on the anomalous trajectories themselves. To address these challenges, our methods exploit contrastive learning techniques and trajectory semantics to capture the patterns underlying the driving scenarios for effective anomaly detection under supervised and unsupervised settings, respectively. We conduct extensive experiments to demonstrate that our supervised method based on contrastive learning and unsupervised method based on reconstruction with semantic latent space can significantly improve the performance of anomalous trajectory detection in their corresponding settings over various baseline methods. We also demonstrate our methods' generalization ability to detect unseen patterns of anomalies. 
\end{abstract}


\section{Introduction}

Tremendous progress has been made for autonomous driving in recent years. The autonomous driving pipeline typically consists of several modules, such as sensing, perception, prediction, planning, and control. In particular, the prediction module encodes other vehicles' past trajectories along with map context and decodes them into potential future trajectories of surrounding vehicles to facilitate the planning module. Recent works~\cite{liu2021multimodal,liang2020learning,yuan2021agentformer,jiao2022tae} have developed various deep learning-based models for trajectory prediction and achieved great performance in terms of the average error between predicted trajectories and ground truth. However, only improving average performance is not enough for autonomous driving systems, where system robustness, safety and security are critical. 

Due to the complexity of real traffic situations and limited coverage of training data, the trajectory prediction task suffers from the ``long-tail'' scenarios by nature. The work in~\cite{zhang2022adversarial} further demonstrates that state-of-the-art trajectory prediction models can be significantly misled by natural-looking but carefully-crafted past trajectory of a certain surrounding vehicle, and discusses several defense methods such as smoothing and SVM-based detection. \cite{jiao2022semi} shows that adversarial training techniques can mitigate the effect of adversarial trajectories.  However, few works focus on advanced \emph{online anomaly detection methods for vehicle trajectories}. We believe that it is crucial to detect anomalous trajectories and scenarios in the prediction stage during runtime, as online anomalous trajectory detection will not only help monitor the prediction module but also enhance the safety of downstream modules in planning~\cite{liu2022neural} and control~\cite{jiao2021end,zhu2021safety}. 

In this work, we consider two different settings -- supervised and unsupervised, based on whether we have prior knowledge of patterns of anomalous trajectories during the training. Both scenarios are possible in real road situations, but they may make a significant difference to methods of learning representations. Thus, we focus on detecting various patterns of anomalous vehicle trajectories in both \emph{supervised} and \emph{unsupervised} settings, and investigate what kinds of representations and corresponding learning techniques are most effective for this safety-critical task.  

The representation for anomalous vehicle trajectory detection is more complicated than that for general time-series anomaly detection because that 1) the driving scenarios contain rich contexts such as road maps and interactions between agents and 2) the anomalous or adversarial trajectories may be associated with specific driving behavior that is difficult to model. To tackle these challenges, an ideal anomaly detector should be able to effectively represent the driving scenario at a single sample level and also model the patterns underlying all normal and anomalous trajectories at the distribution level. Therefore, we first apply a state-of-the-art feature extractor based on graph neural networks to represent the trajectories as well as the road contexts, which is trained on a normal trajectory prediction dataset. Based on the extracted feature, we further add an encoder to capture the distribution-level patterns underlying the anomalous and normal trajectories.  In the supervised setting, we add a contrastive-learning-based encoder to separate the two patterns in the representation space. In the unsupervised setting, we introduce semantics of driving behavior to learn a general and effective latent space for anomaly detection in complex scenarios without labels.  

We extensively compare the anomaly detection performance of different representations under various kinds of anomalies and test scenarios and demonstrate that our proposed representations significantly enhance anomalous trajectory detection performance over baseline methods. The contributions of our work are summarized as follows:

\begin{myitemize}

\item We propose a supervised contrastive learning-based method and an unsupervised method with semantics-guided reconstruction for the anomaly detection of vehicle trajectories and demonstrate their effectiveness in different settings.

\item We explore and compare various representations and architectures for anomalous trajectory detection under supervised and unsupervised settings. We evaluate their performances with three metrics in two different datasets.

\item We further demonstrate the algorithms' generalization ability to detect unseen patterns of anomalies and provide a detailed study to analyze the effectiveness of different modules in our methods.


\end{myitemize}

\section{Background}
\subsection{Anomalous and Adversarial Trajectories}
\begin{figure}[htbp]
\centering
\label{fig:attack-sample}
\subfigure[]{
\begin{minipage}[t]{0.45\linewidth}
\centering
\includegraphics[width=1.58in]{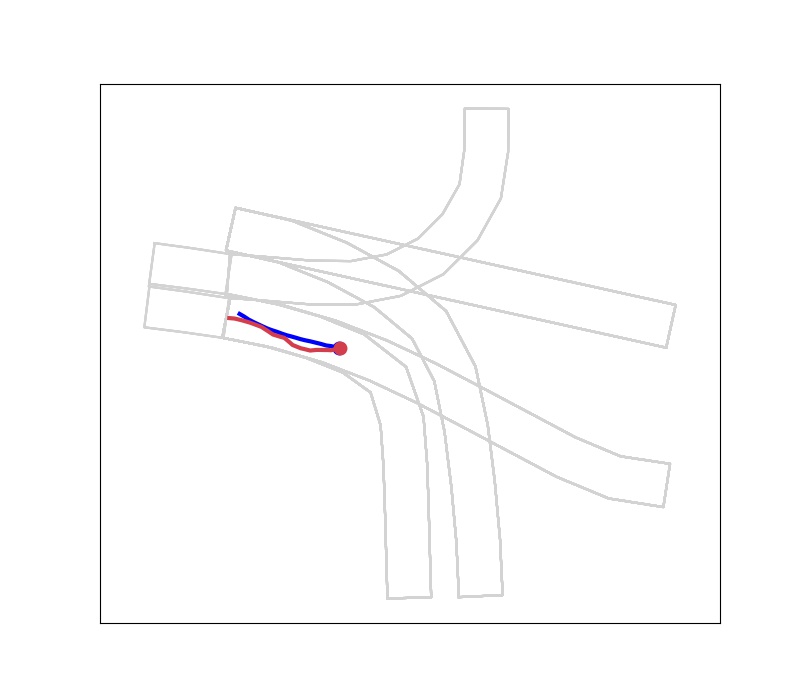}
\\
\includegraphics[width=1.58in]{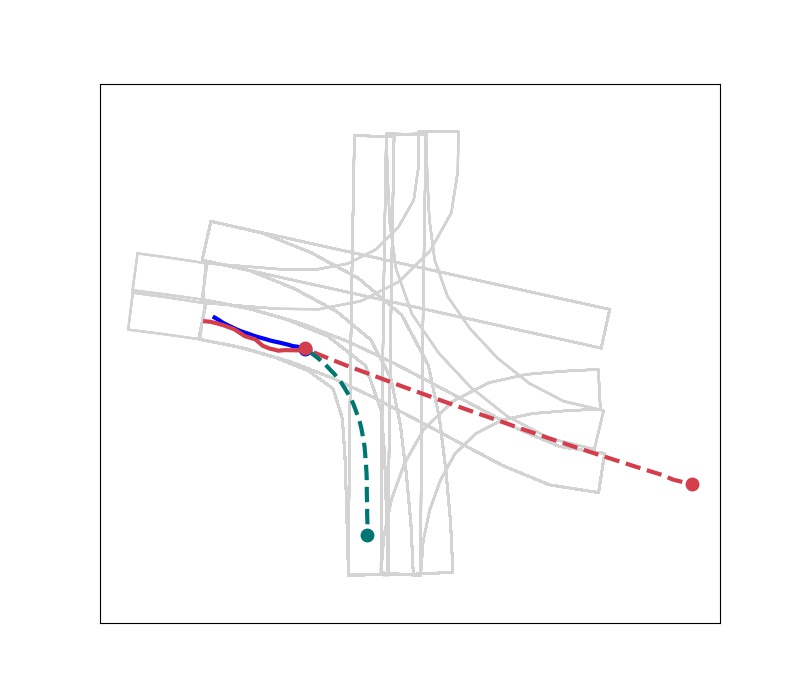}
\end{minipage}%
}%
\subfigure[]{
\begin{minipage}[t]{0.45\linewidth}
\centering
\includegraphics[width=1.58in]{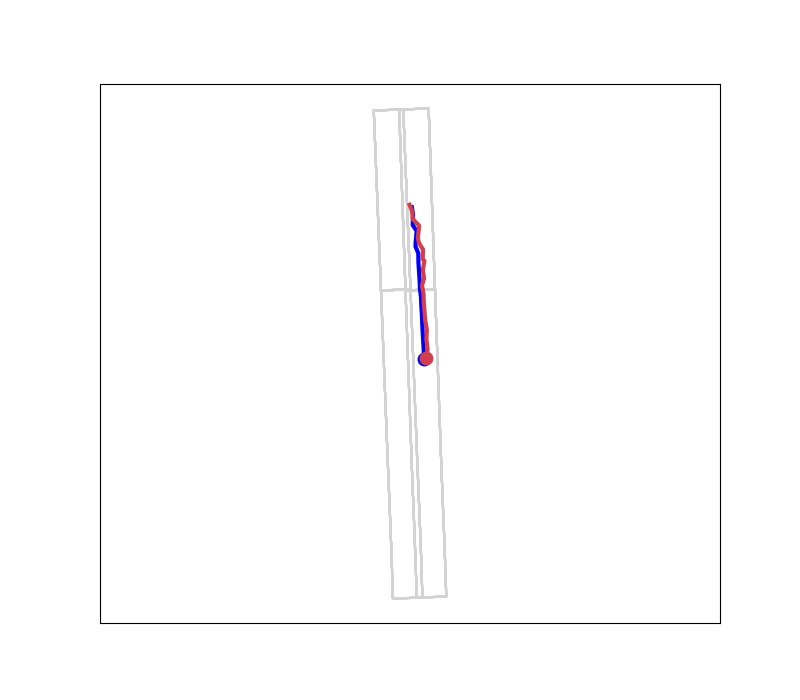}
\\
\includegraphics[width=1.58in]{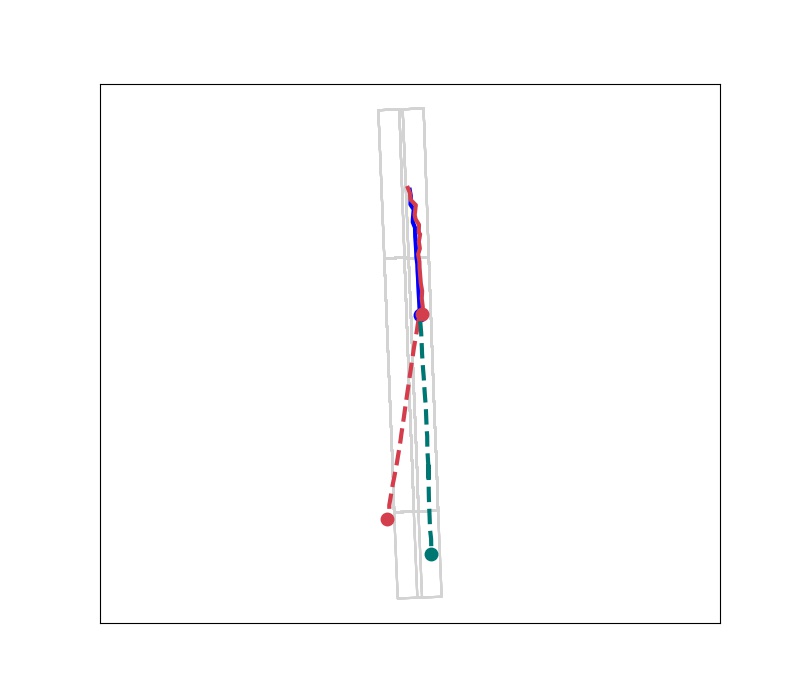}
\end{minipage}%
}%

\centering
\caption{Different patterns and corresponding effects of anomalous trajectories. The figures in the top row are anomalous (red line) and benign (blue line) input history trajectories for prediction and they look very close to human eyes.
The figures in the bottom row show the corresponding affected future trajectory prediction (red dashed line) and the ground truth trajectory in the normal scenario (green dashed line). The differences between the two are clearly visible, showing the great influence caused by anomalous trajectories on prediction modules. Figure (a) is the random anomalous trajectory, which will randomly lead to maximum average deviation. Figure (b) is the lateral directional anomaly, which mainly leads the vehicle to deviate to the left or right.} 
\label{fig:adv_trajs}
\end{figure}

Recent work~\cite{zhang2022adversarial} shows that the trajectory prediction module in autonomous driving pipelines can be easily misled by adversarial (history) trajectories of a surrounding vehicle. 
In a white-box setting, an anomalous trajectory is optimized with Projected Gradient Decent (PGD)~\cite{madry2017towards}. There are different patterns of anomalous trajectories -- random anomalies and directional anomalies. As shown in Fig.~\ref{fig:adv_trajs}, both kinds of anomalous trajectories can effectively interfere with the prediction module and may lead to dangerous scenarios. The random anomaly is a generated trajectory that maximizes the average of the root mean squared error between the predicted and the ground-truth trajectory waypoints. The directional anomalous trajectory is to deliberately mislead the prediction of the surrounding vehicle's future trajectory to a wrong direction. In this work, we apply the lateral directional anomalous as a targeted anomalous pattern and the random attack as a general anomalous pattern, so that we can evaluate the anomaly detection algorithm comprehensively and study their generalization ability to previously-unseen patterns of anomalous trajectories. The detailed metrics for the optimization of targeted attacks are shown in Eq.~\eqref{eq:attack_target}:
\begin{equation}
D(\alpha, R)=\left(p_{\alpha}-s_{\alpha}\right)^{T} \cdot R\left(s_{\alpha+1}, s_{\alpha}\right), \label{eq:attack_target}
\end{equation}

\noindent where $\alpha$ denotes the time frame, $R$ is a function to generate the unit vector to a specific direction (lateral direction in our setting), and $p$ and $s$ are vectors denoting predicted and ground-truth vehicle locations, respectively.

\label{sec:2.1}

\subsection{Anomaly Detection} 
Anomaly detection refers to the problem of finding patterns in data that do not conform to expected behavior\cite{chandola2009anomaly}.  Supervised approaches, unsupervised approaches, and semi-supervised approaches are applied to anomaly detection in different scenarios. 

Supervised approaches generally have better performance on classification tasks, but they require prior knowledge of both normal and anomalous samples.
KNN-based methods \cite{ramaswamy2000efficient,doshi2021online} capture nominal data patterns from the local interaction of nominal data points, and anomalous instances are expected to lie further away from nominal data patterns. The support vector machine (SVM) and neural networks are commonly used to project the input to a feature space and then detect the anomalies from normal data. Some other methods, such as Bayesian networks\cite{mascaro2014anomaly} and inverse reinforcement learning\cite{li2022outlier}, are also effective in supervised anomaly detection.

When labels of anomalies are limited or even unavailable, we have to utilize semi-supervised \cite{ruff2019deep} methods and unsupervised methods for the anomaly detection tasks. Reconstruction-based methods assume that anomalies are not compressible and thus cannot be reconstructed from low-dimensional projections\cite{li2021deep}. In recent works, deep generative models, such as variational autoencoder (VAE)\cite{wiederer2022anomaly,tang2020integrating,8279425}, Generative Adversarial Networks\cite{li2022detecting,schlegl2019f} and adversarial autoencoder\cite{pidhorskyi2018generative}, are introduced to perform reconstruction-based anomaly detection. One-class classification methods including one-class SVM (OC-SVM)\cite{erfani2016high,amer2013enhancing} and one-class neural network (OCNN)\cite{chalapathy2018anomaly} are designed to learn a discriminative boundary surrounding the normal samples.

\subsection{Contrastive Learning}
Contrastive learning \cite{hadsell2006dimensionality} learns representations by contrasting positive pairs against negative pairs. Generally, the augmented versions of the original samples are regarded as positive pairs, and a memory bank is used to stabilize the learning process. Recent works show that contrastive learning techniques can benefit representation learning significantly and there are also some advances in enhancing anomaly detection by utilizing the idea of contrastive learning. For instance, \cite{yue2022ts2vec} proposes an unsupervised method TS2Vec for learning representations of time series. The TS2Vec method captures the contextual representation by leveraging both instance-wise and temporal contrastive loss, and the method shows great performance in time-series anomaly detection. Under a supervised setting, \cite{kopuklu2021driver} demonstrates that the intermediate features of anomaly and normal data can be considered as negative pairs and help learn an effective representation based on contrast.


\subsection{Adversarial Autoencoder}
The variational autoencoder~(VAE)~\cite{kingma2013auto} provides a principled method for jointly learning deep latent-variable models and corresponding inference models using stochastic gradient descent~\cite{kingma2019introduction}, which is commonly used to generate samples in the target space from pre-defined latent distribution. Training a VAE model consists of two kinds of loss: regularization and reconstruction. The regularization is aimed to encode the input as certain distributions over the latent space using Kullback-Leibler (KL) divergence, while the reconstruction is to decode the latent variables to the target or original space.  
In contrast to VAE that uses KL divergence and evidence lower bound, adversarial autoencoder (AAE)~\cite{makhzani2015adversarial} uses adversarial learning to impose a specific distribution on the latent variables, making itself superior to VAE in terms of imposing complicated distributions and shaping the latent space. In our work, we utilize an AAE architecture to model the semantics in the driving scenario.

\section{Our Methods}

\subsection{Feature Extractor}

The anomalous trajectory detection task is more complex compared to general time-series anomaly detection because anomalous trajectory detection highly depends on rich contexts, such as the road map and the behavior of surrounding vehicles. We first feed the map information and the trajectories of vehicles into a feature extractor. Similar to \cite{liang2020learning}, we apply a one-dimensional convolutional network to model history trajectories and utilize graph-neural networks to represent map contexts and interactions between agents. In the training process, we first train a feature extractor in a trajectory prediction pipeline and fix the extractor in the anomaly detection task.

\subsection{Supervised Contrastive Learning-based Method} 

\begin{figure*}[htb]
    \centering
    \includegraphics[width=2\columnwidth]{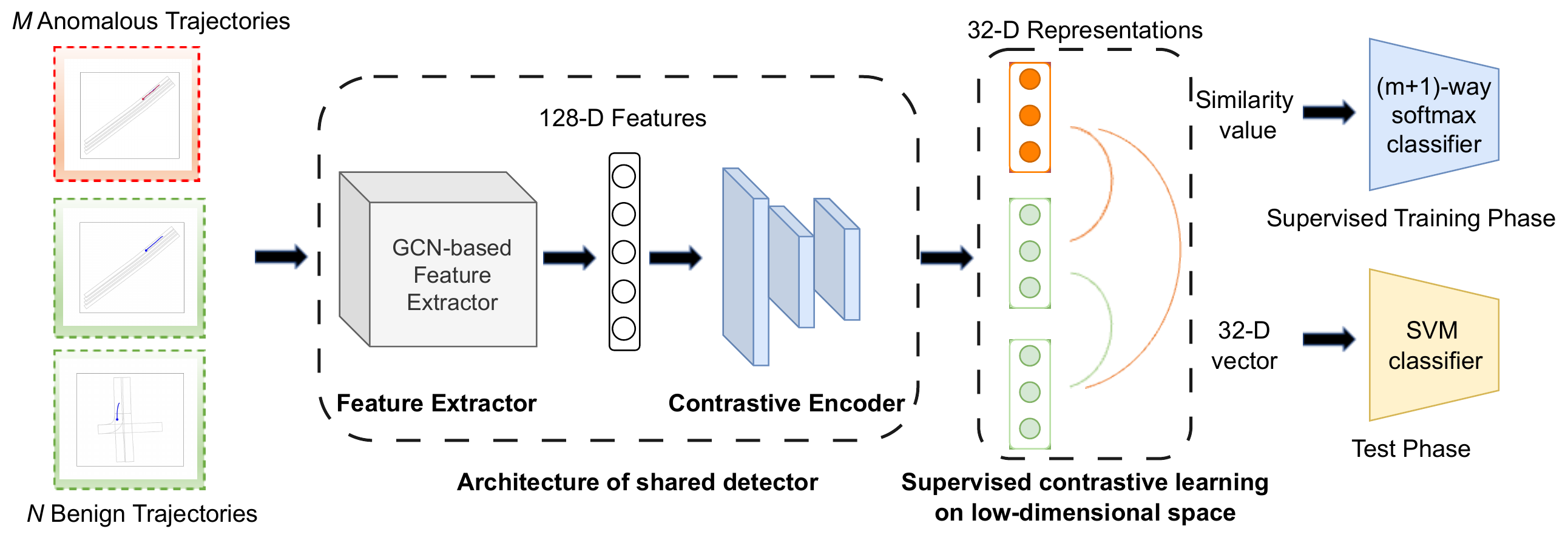}
    \caption{Our supervised contrastive learning-based anomalous trajectory detector. The features of normal and anomalous scenarios are regarded as negative pairs (red ones) and the normal scenarios form positive pairs (green ones).}
    \label{fig:cl_pipeline}
\end{figure*}
As shown in Fig.~\ref{fig:cl_pipeline}, after the feature extractor outputs a representation combining vehicle trajectories and contexts, we further develop a contrastive learning (CL) based encoder to obtain a compact representation for anomalous trajectory detection. Different from instance-wise contrastive learning, the proposed method compares the patterns of two different classes.  This CL-based method is considered supervised because we build the negative pairs by contrasting the anomaly and normal data.  The CL-based encoder is designed to maximize the similarity between benign scenarios and minimize the similarity between benign and anomalous scenarios. Finally, a simple binary classifier based on the encoded representations will generate the decision on whether the input is an anomalous scenario or not.

In every training mini-batch, we have $M$ scenarios containing anomalous trajectories and $N$ normal scenarios, so that we have $N(N-1)$ positive pairs and $MN$ negative pairs, as demonstrated in Fig.~\ref{fig:cl_pipeline}. We use the inner product of two vectors to measure the cosine similarity between encoded features and we set $\tau$ to control the concentration of samples' distribution\cite{hinton2015distilling}.  More negative pairs will generally improve 
 the performance of the learned representation, but it is difficult to calculate and optimize such a large model with $MN$-way softmax vector. To fully utilize the labeled data and keep the model efficient, we apply the idea of Noise Contrastive Estimation (NCE)\cite{gutmann2010noise} to the optimization. We have an $(M+1)$-way softmax classifier (one way for a certain positive pair and $M$ ways for negative pairs) to learn a 32-dimensional representation and the loss function is shown in Eq.~\eqref{cl_eq1}, where $\boldsymbol{s_n}$ and $\boldsymbol{s_a}$ stand for the feature vectors of normal and anomalous scenarios, respectively. Then we define the overall loss function $\mathcal{L}$ for a mini-batch, as shown in Eq.~\eqref{cl_eq2}:

\begin{equation}
\mathcal{L}_{i j}=-\log \frac{\exp \left(\boldsymbol{s}_{\boldsymbol{n} i}^{\mathrm{T}} \boldsymbol{s}_{\boldsymbol{n}j} / \tau\right)}{\exp \left(\boldsymbol{s}_{\boldsymbol{n} i}^{\mathrm{T}} \boldsymbol{s}_{\boldsymbol{n}j} / \tau\right)+\sum_{m=1}^M \exp \left(\boldsymbol{s}_{\boldsymbol{n} i}{ }^{\mathrm{T}} \boldsymbol{s}_{\boldsymbol{am}} / \tau\right)},\label{cl_eq1}
\end{equation}

\begin{equation}
\mathcal{L}=\frac{1}{N(N-1)} \sum_{i=1}^N \sum_{j=1}^N \mathds{1}_{j \neq i} \mathcal{L}_{i j}.\label{cl_eq2}
\end{equation}

During the test time, we add a classifier after the CL-based encoder to produce the detection result based on the 32-dimensional vector. It is feasible to set a threshold of the distance between test samples and the average normal vector to distinguish anomaly from normal data, but in this work, we apply an SVM classifier for all kinds of representations so that we can compare their results fairly. The overall pipeline is demonstrated in Algorithm ~\ref{alg:pipeline}.

\begin{algorithm}[htbp]
\caption{Supervised Contrastive Learning Method}
\label{alg:pipeline}
\begin{algorithmic}[1]
\STATE\textbf{Initialize:} feature extractor $F$, CL-based encoder $E$, Cosine similarity $S_c$ Softmax classifier $C$, and SVM-detector $D$.

\STATE\textbf{Input:} past trajectories $t$ and map graph $g$.

\FOR{each mini-batch}

\STATE 128-D Features $x$ = $F(t,g)$.
\STATE 32-D latent vectors $z$ = $E(x)$.
\STATE N benign trajectories and M anomalous trajectories generate N(N-1) positive pairs and MN negative pairs in the latent space.
\STATE Similarity scores of pairs $s$ = $S_c(z)$.
\FOR{each positive pair ($i,j$)}
\STATE Calculate NCE softmax loss $L_{ij}(s)$ as in~\eqref{cl_eq1}.
\ENDFOR
\STATE Update the encoder $E$ by the CL loss as in~\eqref{cl_eq2}.
\ENDFOR
\STATE Based on the learned 32-D representation, train an SVM classifier for online anomaly detection.
\end{algorithmic}
\end{algorithm}

\subsection{Unsupervised Method using Reconstruction with Semantic Latent Space}
In real road scenes, it may be challenging for us to get valid negative labels for training due to the difficulty of obtaining prior knowledge of surrounding vehicles' potential anomalous trajectories, which motivates us to explore unsupervised detection algorithms. The unsupervised methods are aimed to learn the representation underlying normal driving scenarios and then detect unseen patterns of anomalous trajectories at runtime. Most previous works directly use VAE-based reconstruction, one-class SVM, or contrastive learning (unsupervised) to detect anomalies for time-series data. For vehicle trajectories, however, we can utilize more contexts and domain knowledge to enhance unsupervised anomaly detection. In this work, we propose an unsupervised detection method based on the adversarial autoencoder architecture and semantics modeling in the latent space.   
The encoder takes 128-dimensional features from the extractor as inputs and projects them into a low-dimensional latent space that is divided into three separate parts -- a three-dimensional vector representing lateral intention, a one-dimensional vector representing longitudinal aggressiveness, and a six-dimensional remaining latent vector. Here, we introduce domain knowledge into latent space modeling. We apply \emph{time headway} to extract the longitudinal feature, which measures the time difference between two successive vehicles when crossing a given point. We assume that the time headway follows a log-normal distribution, based on the statistics in urban transportation systems~\cite{jiao2022tae,ha2012time}. The lateral intention is modeled by three simple but reasonable classes that follow categorical distribution: moving forward, turning/changing lanes to the left, and turning/changing lanes to the right. All this semantic information can be collected from benign input trajectory and no knowledge of anomaly is required. For the remaining variables in the latent space, we assume that they follow Gaussian distributions. 

To optimize the latent space, we conduct a two-fold modeling in both overall distributions and semantics of a single vehicle's trajectories. We apply the \emph{adversarial autoencoder} architecture to regularize these distributions of the latent space. Specifically, for each latent vector, a discriminator is trained to distinguish the generated latent vector from the sample in real targeted distribution (log-normal, categorical, or Gaussian). At the same time, we use behavior information such as values of time headway and lateral intention to further render the latent vectors with specific semantics. Thereby, the model can further disentangle the latent space and embed domain knowledge into it. The semantic representation will benefit both unsupervised and supervised anomalous trajectory detection. The loss for semantic latent space modeling is shown below in Eq.~\eqref{eq:semi-loss}:
\begin{equation}
Loss_{sem}(z,g) =  -\sum_{i=1}^3g_{intent}\log z_{intent} + (g_{agg} - z_{agg})^2, \label{eq:semi-loss}
\end{equation}
where $z$ represents the predicted semantic vectors and $g$ represents the reference collected from the input trajectory.

\begin{figure*}[!ht]
    \centering
    \includegraphics[width=2\columnwidth]{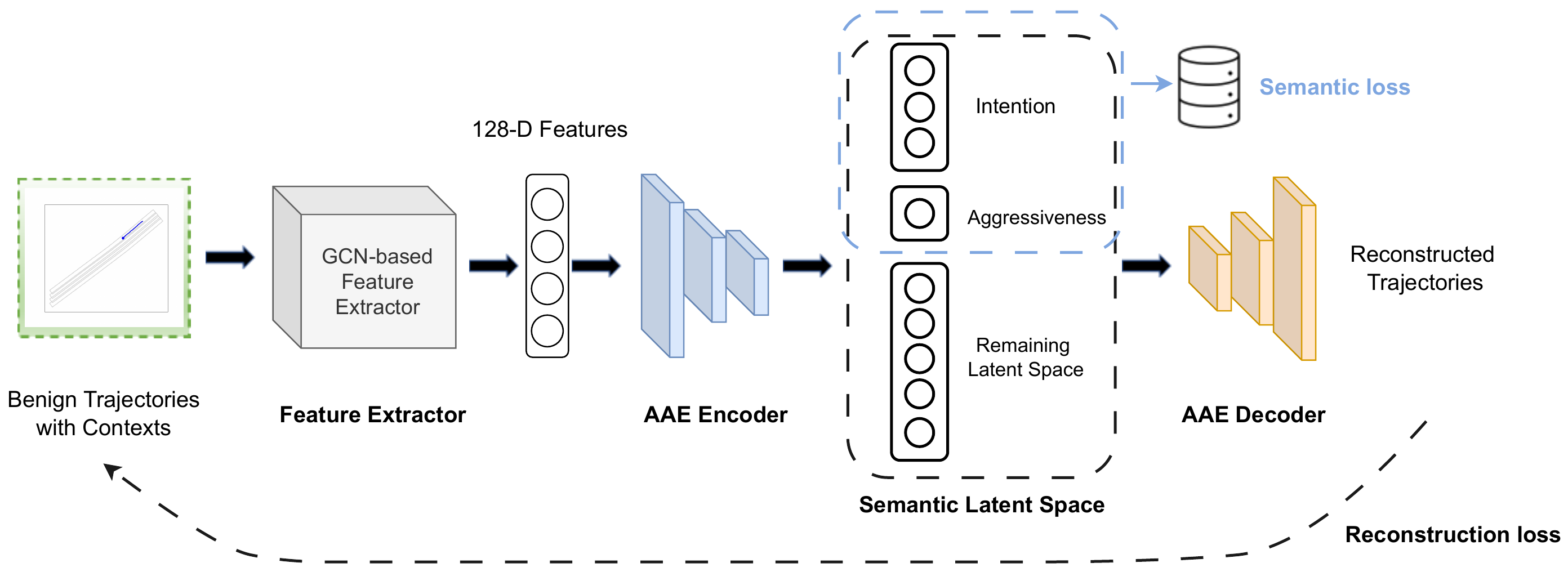}
    \caption{Our unsupervised reconstruction method based on semantic latent space. The latent space contains three kinds of vectors -- intention, aggressiveness, and the remaining vectors. The latent vectors are modeled by semantic labels and target distributions in the adversarial autoencoder.}
    \label{fig:recon_pipeline}
\end{figure*}

The overall pipeline for our unsupervised method is shown in Fig.~\ref{fig:recon_pipeline}. In addition to the semantic latent space modeling, we use a decoder to reconstruct the input trajectories with a smooth L1 loss as shown below in Eq.~\eqref{eq:l1-smooth}: 

\begin{equation}
{Loss}_{recon}\left(y_{i}, \hat{y}_{i}\right)= \begin{cases}0.5\left(y_{i}-\hat{y}_{i}\right)^{2} & \text { if }\left\|y_{i}-\hat{y}_{i}\right\|<1, \\ \left\|y_{i}-\hat{y}_{i}\right\|-0.5 & \text { otherwise },\end{cases} \label{eq:l1-smooth}
\end{equation}
where $y$ and $\hat{y}$ represent the input trajectories and reconstructed trajectories, respectively. Both the encoder and the decoder will be optimized by the reconstruction loss. The overall optimization pipeline is shown in Algorithm~\ref{alg:semantics}.

Note that in an unsupervised setting, we use the error between the input trajectory and reconstructed trajectory as a signal for anomaly detection. We consider the input as an anomalous trajectory if the reconstruction error is larger than a threshold. The learned representation can also be used in a supervised setting by adding common binary classifiers after the latent space.

\begin{algorithm}[htbp]
\caption{Unsupervised Semantic Reconstruction Method}
\label{alg:semantics}
\begin{algorithmic}[1]
\STATE\textbf{Initialize:} feature extractor $F$, AAE encoder $E$, decoder $R$, discriminator $D_{i}$, target distribution $p_{i}$, $i$ = 1,2,3.

\STATE\textbf{Input:} past trajectories $t$ and map graph $g$.

\FOR{each batch}
\STATE Features $x$ = $F(t,m)$.
\STATE Let latent vectors $z$ = $E(x)$.
\STATE Sample $s_i$ from target distribution $p_{i}$ and calculate  $D_{i}(z_i)$ and $D_{i}(s_i)$.
\STATE Update $E$ and $D_i$ by discrimination loss and generation loss as in~\cite{makhzani2015adversarial}. 
\STATE Calculate the true value for intention and aggressiveness, respectively.
\STATE Update $E$ by semantic loss $Loss_{sem}$ as in~\eqref{eq:semi-loss}.
\STATE Concatenate the latent vectors and feed them to the detector $R$ $\hat y$ = $R(z)$.
\STATE Update $G$, $R$ by reconstruction loss $Loss_{recon}(y,\hat y)$ as in~\eqref{eq:l1-smooth}.
\ENDFOR

\end{algorithmic}
\end{algorithm}

\section{Experiments}

In this section, we present the anomaly detection results of six methods under two patterns of anomalous scenarios. Each method is tested in two commonly-used datasets with three different metrics so as to comprehensively evaluate the performance, especially under imbalanced data distribution. The results show that the representation learned by our supervised contrastive learning can significantly improve detection performance. Moreover, the semantic latent space we construct can effectively model the context and explicitly encode driving behavior, enhancing anomaly detection in both supervised and unsupervised settings. We conduct a study to show to what extent the `semantics' and `contrast' can benefit the representation learning for anomaly trajectory detection. In addition, we evaluate the generalization ability of learned representations, which is critical for detecting unseen patterns of anomalies. 

\subsection{Experiment Setup}

\subsubsection{Data Collection}
We conduct experiments with both random and directional anomalous trajectories on two datasets: Argoverse 1~\cite{chang2019argoverse} and Argoverse 2~\cite{wilson2023argoverse}. The Argoverse 1 motion forecasting dataset has more than 30K driving scenarios collected in Miami and Pittsburgh, while Argoverse 2 collects longer and more complicated driving scenarios in six cities. Each scenario used in this work consists of a road graph and trajectories of multiple agents.
The history trajectories are 20 waypoints collected in the past 2 seconds. 

To collect the anomalous trajectories, we apply the attack methods mentioned in Sec.~\ref{sec:2.1} to generate different patterns of anomalies. For lateral directional anomalous trajectories, we consider past trajectories that can lead to a prediction error of more than 1.5 meters in a lateral direction as anomalies. For random anomalous trajectories, the threshold is set as 5-meter average displacement error (ADE).

\subsubsection{Evaluation Metrics}
We utilize three different metrics to evaluate the performance of anomaly detection approaches -- ROC AUC (area under the receiver operating characteristic curve), PR AUC (area under the precision-recall curve), and F1 score. The ROC AUC is a general metric to evaluate the binary classification ability at all classification thresholds, but it can be overly optimistic on severely imbalanced classification problems. For the imbalance dataset in anomaly detection, the PR AUC is a more powerful metric because both precision and recall are focused on the positive (anomaly) class and are unconcerned with the majority class.  The F1 score is the harmonic mean of precision and recall. In the anomaly detection task, the recall (detected anomalies over all anomalies) is expected to be high. Thus we find the point where recall is fixed as 0.8 and calculate its corresponding F1 score. 

\subsection{Effectiveness of Our CL-based Supervised Method}

In the supervised setting, the labels of anomalous driving scenarios are available. We use SVM as a fixed classifier to compare the results of different learned representations. The naive SVM method builds an SVM directly on the acceleration series of the input trajectories. For the methods with semantic latent space and contrastive learning encoder, we use the 128-dimensional feature produced by the feature extractor as input. The results in Tables~\ref{tab:sup_argo1} and~\ref{tab:sup_argo2} show that \textbf{our contrastive learning-based supervised method greatly outperforms other supervised methods in both directional and random anomaly patterns}. Compared to the method using semantic latent space (`Semantics + SVM'), our CL-based supervised method (`Sup-CL + SVM') can effectively model the distribution of normal and abnormal trajectories and separate them in the CL-based representation space, making it easy for a simple SVM classifier to detect anomalies. The results of naive SVM (`Naive SVM') demonstrate that it is difficult to directly distinguish anomaly from normal data in the trajectory space, even with enough labels.

\begin{table}[h]

\centering
\caption{Results of supervised anomaly detection methods for the Argoverse 1 dataset.}   
\resizebox{\columnwidth}{!}{
\label{tab:sup_argo1}

\begin{tabular}{|c|c|c|c|c|}
\hline
Methods & Anomaly Pattern   & F1 score & ROC AUC &PR AUC   \\ \hline

Naive SVM~\cite{zhang2022adversarial} &Random& $0.51$ &$0.63$ &  $0.43$ \\ \hline
Naive SVM~\cite{zhang2022adversarial} &Lateral& $0.49$ &$0.75$  &$0.51$\\ \hline
Semantics + SVM  &Random & $0.68$  &$0.85$  &$0.74$ \\ \hline
Semantics + SVM &Lateral & $0.84$  &$0.96$&   $0.92$\\ \hline
\textbf{Sup-CL + SVM} & \textbf{Random}  & $\textbf{0.81}$  &$\textbf{0.93}$ &$\textbf{0.86}$ \\ \hline
\textbf{Sup-CL + SVM} & \textbf{Lateral} & $\textbf{0.87}$ & $\textbf{0.98}$ &$\textbf{0.96}$\\ \hline

\end{tabular}}
\end{table}

\begin{table}[h]

\centering
\caption{Results of supervised anomaly detection methods for the Argoverse 2 dataset.}   
\resizebox{\columnwidth}{!}{
\label{tab:sup_argo2}

\begin{tabular}{|c|c|c|c|c|}
\hline
Methods & Anomaly Pattern   & F1 score & ROC AUC &PR AUC   \\ \hline

Naive SVM~\cite{zhang2022adversarial} &Random& $0.42$ &$0.59$ &$0.31$\\ \hline
Naive SVM~\cite{zhang2022adversarial} &Lateral& $0.47$ &$0.64$ &$0.40$  \\ \hline
Semantics + SVM  &Random & $0.62$  &$0.84$ &$0.65$  \\ \hline
Semantics + SVM &Lateral & $0.84$  &$0.95$ &$0.89$   \\ \hline
\textbf{Sup-CL + SVM} & \textbf{Random}  & $\textbf{0.80}$  &$\textbf{0.94}$  & $\textbf{0.87}$\\ \hline
\textbf{Sup-CL + SVM} & \textbf{Lateral} & $\textbf{0.99}$ &$\textbf{0.98}$ &$\textbf{0.99}$\\ \hline

\end{tabular}}
\end{table}

\begin{figure*}[htbp]

\centering

\subfigure[]{
\begin{minipage}[t]{0.45\linewidth}
\centering
\includegraphics[width=3in]{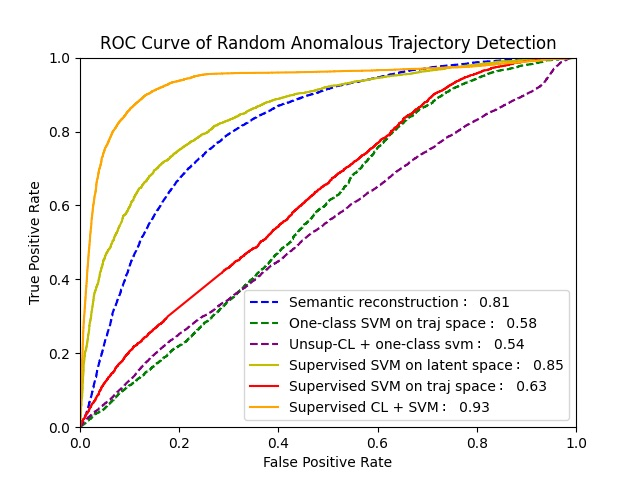}
\end{minipage}%
}%
\subfigure[]{
\begin{minipage}[t]{0.45\linewidth}
\centering
\includegraphics[width=3in]{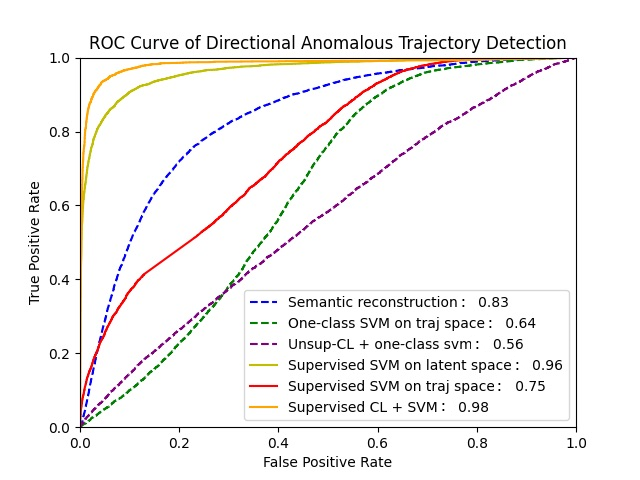}
\end{minipage}%
}%

\caption{ROC curves of both supervised (solid lines) and unsupervised (dashed lines) anomalous trajectory detection approaches The AUC values are shown in legends. The left figure shows the results of detecting \textbf{random} anomalous trajectories when all supervised methods are trained on the \textbf{random} anomalies. The right figure shows the results of detecting \textbf{lateral directional} anomalous trajectories when all supervised methods are trained on the \textbf{directional} anomalies.}
\label{fig:same_anomaly}
\vspace{-0.5cm}
\end{figure*}

\begin{figure*}[htbp]
\label{}
\centering

\subfigure[]{
\begin{minipage}[t]{0.45\linewidth}
\centering
\includegraphics[width=3in]{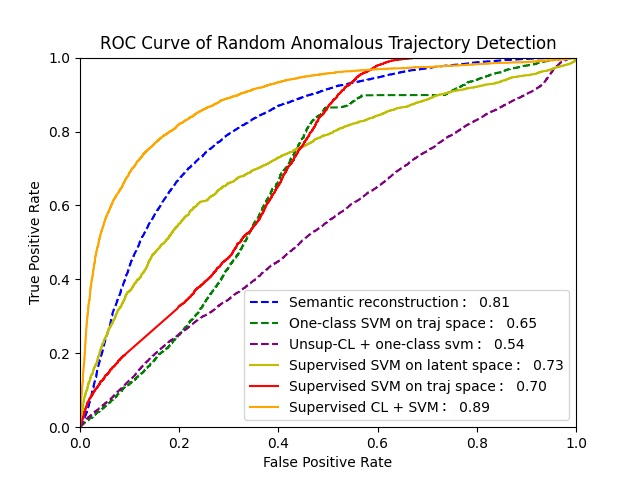}
\end{minipage}%
}%
\subfigure[]{
\begin{minipage}[t]{0.45\linewidth}
\centering
\includegraphics[width=3in]{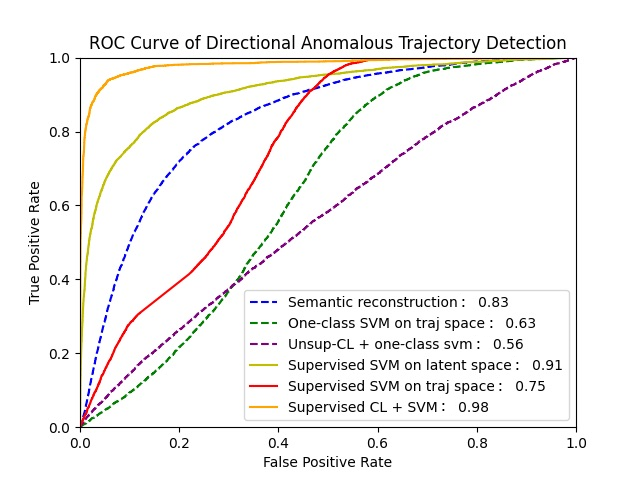}
\end{minipage}%
}%

\caption{ROC curves of both supervised (solid lines) and unsupervised (dashed lines) anomalous trajectory detection approaches under \textbf{unseen} patterns of anomalies. The AUC values are shown in legends. The left figure shows the results of detecting \textbf{random} anomalous trajectories when all supervised methods are trained on the \textbf{lateral directional} anomalies. The right figure shows the results of detecting \textbf{lateral directional} anomalous trajectories when all supervised methods are trained on the \textbf{random} anomalies.}
\label{fig:diff_anomaly}
\vspace{-0.5cm}
\end{figure*}

\subsection{Effectiveness of Our Unsupervised Method with Semantic Reconstruction}

In the unsupervised case, without prior knowledge of anomaly patterns, it is more difficult to learn an informative representation. We utilize the one-class SVM on trajectory space~\cite{erfani2016high} and a state-of-the-art unsupervised contrastive learning method -- TS2Vec~\cite{yue2022ts2vec} as baselines. As shown in Tables~\ref{tab:unsup_argo1} and~\ref{tab:unsup_argo2}, the one-class SVM on trajectory space method (`OC-SVM') and the unsupervised contrastive learning method (`Unsup-CL + OC-SVM') have relatively poor performance and can hardly detect the anomalous trajectories. \textbf{Our unsupervised method with semantic reconstruction (`Semantic Recon') has much better performance on every metric over other unsupervised methods.} In addition, when detecting random anomalous trajectories, we find that our unsupervised method has close performance to its corresponding supervised version, which further demonstrates the effectiveness of our semantic latent representation. The ROC curves of all supervised and unsupervised methods are shown in Fig.~\ref{fig:same_anomaly}.

\begin{table}[h]
\centering
\caption{Results of unsupervised anomaly detection methods for the Argoverse 1 dataset.}   
\resizebox{\columnwidth}{!}{
\label{tab:unsup_argo1}

\begin{tabular}{|c|c|c|c|c|}
\hline
Methods & Anomaly Pattern   & F1 score & ROC AUC &PR AUC   \\ \hline

OC-SVM~\cite{erfani2016high} &Random& $0.51$ &$0.58$  & $0.36$\\ \hline
OC-SVM~\cite{erfani2016high} &Lateral& $0.47$ &$0.64$  &$0.30$\\ \hline
Unsup-CL + OC-SVM~\cite{yue2022ts2vec} &Random & $0.41$  &$0.54$  & $0.30$\\ \hline
Unsup-CL + OC-SVM~\cite{yue2022ts2vec} &Lateral & $0.40$  & $0.56$ & $0.29$ \\ \hline
\textbf{Semantic Recon} & \textbf{Random}  &$\textbf{0.65}$  &$\textbf{0.81}$ & $\textbf{0.61}$\\ \hline
\textbf{Semantic Recon} & \textbf{Lateral} & $\textbf{0.60}$  &$\textbf{0.83}$& $\textbf{0.56}$\\ \hline

\end{tabular}}
\end{table}

\begin{table}[h]

\centering
\caption{Results of unsupervised anomaly detection methods for the Argoverse 2 dataset.}   
\resizebox{\columnwidth}{!}{
\label{tab:unsup_argo2}

\begin{tabular}{|c|c|c|c|c|}
\hline
Methods & Anomaly Pattern   & F1 score & ROC AUC &PR AUC   \\ \hline

OC-SVM~\cite{erfani2016high} &Random& $0.42$ &$0.55$ &  $0.28$\\ \hline
OC-SVM~\cite{erfani2016high} &Lateral& $0.40$ &$0.54$ &$0.26$ \\ \hline
Unsup-CL + OC-SVM~\cite{yue2022ts2vec} &Random & $0.41$  &$0.54$ &  $0.30$ \\ \hline
Unsup-CL + OC-SVM~\cite{yue2022ts2vec} &Lateral & $0.38$  &$0.57$ &$0.27$   \\ \hline
\textbf{Semantic Recon} & \textbf{Random}  & $\textbf{0.53}$ & $\textbf{0.77}$ & $\textbf{0.47}$\\ \hline
\textbf{Semantic Recon} & \textbf{Lateral} &$\textbf{0.53}$ &$\textbf{0.79}$ &$\textbf{0.48}$ \\ \hline

\end{tabular}}
\end{table}

\subsection{Effectiveness of Components in Our Proposed Methods}
We conduct more experiments to study to what extent different components of our methods benefit the overall performance improvement.
In this study, we test all representations in a supervised manner for comparison. We evaluate how much the feature extractor (embedding the contexts) and the following encoder (semantics or CL-based) contribute to the performance, respectively. The `NN + SVM' stands for the method of directly adding an SVM classifier after the GNN-based feature extractor. In Table~\ref{tab:ablation1}, we find that using such representation directly from the feature extractor has a much poorer detection performance than our CL-based representation (`Sup-CL + SVM'). The latent space modeling methods (`Semantics + SVM' and `Naive Latent + SVM') also outperform the pure feature extractor (`NN + SVM'). Moreover, Table~\ref{tab:ablation2}  
reveals that our semantic latent space modeling (`Semantics + SVM') significantly improves the generalization ability to unseen patterns of anomalies when compared to the naive latent space modeling without any semantics (`Naive Latent + SVM') and the pure feature extractor (`NN + SVM').

\begin{table}[h]

\centering
\caption{Effectiveness of components: trained and tested on random anomalies.}   
\resizebox{\columnwidth}{!}{
\label{tab:ablation1}

\begin{tabular}{|c|c|c|c|c|}
\hline

Methods & Anomaly Pattern   & F1 score & ROC AUC &PR AUC   \\ \hline

NN + SVM  &Random & $0.66$  &$0.61$ & $0.68$  \\ \hline

Naive Latent + SVM &Random & $0.68$  &$0.86$  &$0.74$ \\ \hline

Semantics + SVM &Random  & $0.68$  &$0.85$ & $0.74$\\ \hline

Sup-CL + SVM &Random  & $0.81$  &$0.93$  &$0.86$\\ \hline

\end{tabular}}
\end{table}

\begin{table}[h]

\centering
\caption{Effectiveness of components: trained on directional anomalies and tested on random anomalies.}   
\resizebox{\columnwidth}{!}{
\label{tab:ablation2}

\begin{tabular}{|c|c|c|c|c|}
\hline
Methods & Anomaly Pattern   & F1 score & ROC AUC &PR AUC   \\ \hline

NN + SVM  &Random & $0.44$  &$0.60$ & $0.35$  \\ \hline

Naive Latent + SVM &Random & $0.44$  &$0.58$  & $0.45$ \\ \hline

Semantics + SVM  &Random  & $0.65$  &$0.81$ & $0.58$ \\ \hline

Sup-CL + SVM &Random  & $0.70$  &$0.89$  &$0.75$\\ \hline

\end{tabular}}
\end{table}

\subsection{Evaluation on Generalization Ability}

In the supervised setting, one critical aspect is how the representation learned from normal samples and a certain pattern of anomalies can be generalized to other unseen patterns of anomalies. Fig.~\ref{fig:diff_anomaly} shows the results when the supervised methods are trained and tested on different patterns of anomalies. Compared to Fig.~\ref{fig:same_anomaly}, we find that the lateral directional anomalies are relatively easy to detect, even when the models are trained on another kind of anomaly. However, when the models are trained on lateral directional anomalies but tested on the random anomalous trajectories, the performances of supervised methods drop significantly, although the supervised CL-based method is still the best, which reveals overfitting and a lack of ability to generalize. In this setting, the unsupervised reconstruction with semantic latent space even outperforms its corresponding supervised version. Table~\ref{tab:ablation2} further shows our semantics modeling can help in learning a more generalized representation and mitigating overfitting to a certain pattern of anomalies, compared to the naive latent space.  

\section{Conclusions}
In this work, we propose novel contrastive learning-based supervised method and semantic reconstruction-based unsupervised method for anomalous vehicle trajectory detection. We embed both driving contexts and distributions underlying the normal (and anomalous) trajectories into the representation by various methods. Extensive experiments demonstrate that our methods can significantly improve the detection performance over baseline methods in supervise and unsupervised settings, respectively.  
We also demonstrate that our methods have better generalization ability to address unseen attack patterns, which is valuable for practical applications on the road.




\bibliographystyle{IEEEtran}
\bibliography{reference}

\end{document}